\documentstyle[colacl,11pt]{article}
\title{{\bf Introduction to the CoNLL-2001 Shared Task: \\
            Clause Identification}}
\author{
\begin{tabular}{cc}
Erik F. Tjong Kim Sang             & Herv\'e D\'ejean\\
CNTS -- Language Technology Group  & Seminar f\"ur Sprachwissenschaft\\
University of Antwerp              & Universit\"at T\"ubingen\\
{\it erikt@uia.ua.ac.be}           & {\it dejean@sfs.nphil.uni-tuebingen.de}
\end{tabular}
}

\date{\today}

\begin{document}

\maketitle


\vspace*{-0.0cm}
\begin{abstract}
\noindent
We describe the CoNLL-2001 shared task: dividing text into
clauses.
We give background information on the data sets, present a general
overview of the systems that have taken part in the shared task and
briefly discuss their performance.
\end{abstract}

\section{Introduction}

The CoNLL-2001 shared task aims at discovering clause boundaries with
machine learning methods.
Why clauses? 
Clauses are structures used in applications such as 
Text-to-Speech conversion \cite{ejerhed88}, 
text-alignment \cite{papageorgiou97} and 
machine translation \cite{leffa98}.
Ejerhed \shortcite{ejerhed88} described clauses as a natural structure
above chunks: 

\begin{quote}
It is a hypothesis of the author's current clause-by-clause processing theory,
that a unit corresponding to the basic clause is a stable and easily recognizable surface unit and that
is is also an important partial result and building block in the construction od a richer linguistic
representation that encompasses syntax as well as semantics and discourse structure \cite[page 220]{ejerhed88}
\end{quote}

The goal of this shared task is to evaluate automatic methods,
especially machine learning methods, for finding clause boundaries 
in text.
We have selected a training and test corpus for performing this
evaluation.
The task has been divided in three parts in order to allow basic
machine learning methods to participate in this task by processing the
data in a bottom-up fashion.

\section{Task description}
\label{sec-task}

Defining clause boundaries is not trivial \cite{leffa98}.
In this task, the gold standard clause segmentation is provided by the
Penn Treebank \cite{marcus93}. 
The guidelines of the Penn Treebank describe in detail how sentences
are segmented into clauses \cite{bies95}.
Here is an example of a sentence and its clauses obtained from 
Wall Street Journal section 15 of the Penn Treebank \cite{marcus93}:

\vspace*{\baselineskip}
\noindent
(S Coach them in\\
\hspace*{0.25cm}(S--NOM handling complaints)\\
\hspace*{0.25cm}(SBAR--PRP so that\\
\hspace*{0.50cm}(S they can resolve problems immediately)\\
\hspace*{0.25cm})\\
\hspace*{0.25cm}.\\
)

\vspace*{\baselineskip}
\noindent
The clauses of this sentence have been enclosed between brackets.
A tag next to the open bracket denotes the type of the clause.

In the CoNLL-2001 shared task, the goal is to identify clauses in
text.
Since clauses can be embedded in each other, this task is considerably
more difficult than last year's task, recognizing non-embedded text
chunks.
For that reason, we have disregarded type and function information
of the clauses: every clause has been tagged with S rather
than with an elaborate tag such as SBAR--PRP.
Furthermore, the shared task has been divided in three parts:
identifying clause starts, recognizing clause ends and finding
complete clauses.
The results obtained for the first two parts can be used in the 
third part of the task.

\section{Data and Evaluation}

This CoNLL shared task works with roughly the same
sections of the Penn Treebank as the widely used data set for 
base noun phrase recognition \cite{ramshaw95}:
WSJ sections 15--18 of the Penn Treebank as training material, 
section 20 as development material for tuning the parameter of the
learner and section 21 as test data\footnote{ 
 These clause data sets are available at\\
 http://lcg--www.uia.ac.be/conll2001/clauses/
}.
The data sets contain tokens (words and punctuation marks), information 
about the location of sentence boundaries and information about clause
boundaries. 
Additionally, a part-of-speech (POS) tag and a chunk tag was assigned
to each token by a standard POS tagger \cite{brill94} and a chunking
program \cite{tks2000d}.
We used these POS and chunking tags rather than the Treebank ones in
order to make sure that the performance rates obtained for this data
are realistic estimates for data for which no Treebank tags are
available. 
In the clause segmentation we have only included clauses in the
Treebank which had a label starting with S thus disregarding
clauses with label RRC or FRAG.
All clause labels have been converted to S.

Different schemes for encoding phrase information have been used in
the data:

\begin{itemize}
\item B-X, I-X and O have been used for marking the first word in a
   chunk of type X, a non-initial word in an X chunk and a word
   outside of any chunk, respectively
   (see also Tjong Kim Sang and Buchholz \shortcite{tks2000c}).
\item S, E and X mark a clause start, a clause end and neither a 
   clause start nor a clause end, respectively.
   These tags have been used in the first and second part of the
   shared task.
\item (S*, *S) and * denote a clause start, a clause end and neither a 
   clause start nor a clause end, respectively.
   The first two can be used in combination with each other.
   For example, (S*S) marks a word where a clause starts and ends,
   and *S)S) marks a word where two clauses end.
   These tags are used in the third part of the shared task.
\end{itemize}

\noindent
The first two phrase encodings were inspired by the representation
used by Ramshaw and Marcus \shortcite{ramshaw95}.
Here is an example of the clause encoding schemes:

\begin{center}
\begin{tabular}{rlll}
Coach&S&X&(S* \\
them&X&X&* \\
in&X&X&* \\
handling&S&X&(S* \\
complaints&X&E&*S) \\
so&S&X&(S* \\
that&X&X&* \\
they&S&X&(S* \\
can&X&X&* \\
resolve&X&X&* \\
problems&X&X&* \\
immediately&X&E&*S)S)  \\
.&X&E&*S)
\end{tabular}
\end{center}

\noindent
Three tags can be found next to each word, respectively denoting the
information for the first, second and third part of the shared task.
The goal of this task is to predict the test data segmentation as
well as possible with a model built from the training data.

The performance in this task is measured with three rates.
First, the percentage of detected starts, ends or clauses 
that are correct (precision). 
Second, the percentage of starts, ends or clauses in the data that
were found by the learner (recall).
And third, the F$_{\beta=1}$ rate which is equal to
($\beta^2$+1)*precision*recall / ($\beta^2$*precision+recall)
with $\beta$=1
\cite{vanrijsbergen75}.
The latter rate has been used as the target for 
optimization.

\begin{table}[t]
\begin{center}
\begin{tabular}{|l|c|c|c|l}\cline{2-4}
\multicolumn{1}{l|}{development 1} 
                      & precision & recall & F$_{\beta=1}$ \\\cline{1-4}
Carreras \& M\`ar.    & 95.77\% & 92.08\% & 93.89 \\
Patrick \& Goyal      & 94.84\% & 87.33\% & 90.93 & \hspace*{-0.1cm}*$^2$ \\
Tjong Kim Sang        & 92.94\% & 86.87\% & 89.80 & \hspace*{-0.1cm}* \\
Molina \& Pla         & 90.11\% & 88.80\% & 89.45 & \hspace*{-0.1cm}* \\
D\'ejean              & 94.08\% & 84.59\% & 89.08 \\\cline{1-4}
baseline              & 96.32\% & 38.08\% & 54.58\\\cline{1-4}
\end{tabular}

\vspace*{0.5cm}
\begin{tabular}{|l|c|c|c|l}\cline{2-4}
\multicolumn{1}{l|}{test part 1} 
                      & precision & recall & F$_{\beta=1}$ \\\cline{1-4}
Carreras \& M\`ar.    & 93.96\% & 89.59\% & 91.72 \\
Tjong Kim Sang        & 92.91\% & 85.08\% & 88.82 & \hspace*{-0.1cm}*\\
Molina \& Pla         & 89.54\% & 86.01\% & 87.74 & \hspace*{-0.1cm}*\\
D\'ejean              & 93.76\% & 81.90\% & 87.43 \\
Patrick \& Goyal      & 89.79\% & 84.88\% & 87.27 & \hspace*{-0.1cm}*\\\cline{1-4}
baseline              & 98.44\% & 36.58\% & 53.34\\\cline{1-4}
\end{tabular}

\end{center}
\caption{
The performance of five systems while processing the development
data and the test data for part 1 of the shared task: finding clause
starts. 
The baseline results have been obtained by a system that assumes that
every sentence consists of one clause which contains the complete
sentence. 
} 
\label{tab-task1}
\end{table}

\begin{table}[t]
\begin{center}
\begin{tabular}{|l|c|c|c|l}\cline{2-4}
\multicolumn{1}{l|}{development 2} 
                      & precision & recall & F$_{\beta=1}$ \\\cline{1-4}
Carreras \& M\`ar.    & 91.27\% & 89.00\% & 90.12 \\
Tjong Kim Sang        & 83.80\% & 80.44\% & 82.09 \\
Patrick \& Goyal      & 80.12\% & 83.03\% & 81.55 & \hspace*{-0.1cm}* \\
Molina \& Pla         & 78.65\% & 78.97\% & 78.81 & \hspace*{-0.1cm}* \\
D\'ejean              & 99.28\% & 51.73\% & 68.02 \\\cline{1-4}
baseline              & 96.32\% & 51.86\% & 67.42\\\cline{1-4}
\end{tabular}

\vspace*{0.5cm}
\begin{tabular}{|l|c|c|c|l}\cline{2-4}
\multicolumn{1}{l|}{test part 2} 
                      & precision & recall & F$_{\beta=1}$ \\\cline{1-4}
Carreras \& M\`ar.    & 90.04\% & 88.41\% & 89.22 \\
Tjong Kim Sang        & 84.72\% & 79.96\% & 82.28 \\
Patrick \& Goyal      & 80.11\% & 83.47\% & 81.76 & \hspace*{-0.1cm}* \\
Molina \& Pla         & 79.57\% & 77.68\% & 78.61 & \hspace*{-0.1cm}* \\
D\'ejean              & 99.28\% & 48.90\% & 65.47 \\\cline{1-4}
baseline              & 98.44\% & 48.90\% & 65.34\\\cline{1-4}
\end{tabular}

\end{center}
\caption{
The performance of five systems while processing the development
data and the test data for part 2 of the shared task: identifying
clause ends.
The baseline results have been obtained by a system that assumes that
every sentence consists of one clause which contains the complete
sentence. 
} 
\label{tab-task2}
\end{table}

\begin{table}[t]
\begin{center}
\begin{tabular}{|l|c|c|c|l}\cline{2-4}
\multicolumn{1}{l|}{development 3} 
                      & precision & recall & F$_{\beta=1}$ \\\cline{1-4}
Carreras \& M\`ar.    & 87.18\% & 82.48\% & 84.77 \\
Patrick \& Goyal      & 78.19\% & 67.63\% & 72.53 & \hspace*{-0.1cm}* \\
Molina \& Pla         & 70.98\% & 72.31\% & 71.64 & \hspace*{-0.1cm}* \\
Tjong Kim Sang        & 76.54\% & 67.20\% & 71.57 & \hspace*{-0.1cm}* \\
D\'ejean              & 73.93\% & 62.44\% & 67.70 \\
Hammerton             & 59.85\% & 55.56\% & 57.62 \\\cline{1-4}
baseline              & 96.32\% & 35.77\% & 52.17\\\cline{1-4}
\end{tabular}

\vspace*{0.5cm}
\begin{tabular}{|l|c|c|c|l}\cline{2-4}
\multicolumn{1}{l|}{test part 3} 
                      & precision & recall & F$_{\beta=1}$ \\\cline{1-4}
Carreras \& M\`ar.    & 84.82\% & 73.28\% & 78.63 \\
Molina \& Pla         & 70.89\% & 65.57\% & 68.12 & \hspace*{-0.1cm}* \\
Tjong Kim Sang        & 76.91\% & 60.61\% & 67.79 & \hspace*{-0.1cm}*\\
Patrick \& Goyal      & 73.75\% & 60.00\% & 66.17 & \hspace*{-0.1cm}*\\
D\'ejean              & 72.56\% & 54.55\% & 62.77 \\
Hammerton             & 55.81\% & 45.99\% & 50.42 \\\cline{1-4}
baseline              & 98.44\% & 31.48\% & 47.71\\\cline{1-4}
\end{tabular}

\end{center}
\caption{
The performance of the six systems while processing the development
data and the test data for part 3 of the shared task: recognizing
complete clauses.
The baseline results have been obtained by a system that assumes that
every sentence consists of one clause which contains the complete
sentence. 
} 
\label{tab-task3}
\end{table}

\section{Results}

Six systems have participated in the shared task.
Two of them used boosting and the others used techniques which were
connectionist, memory-based, statistical and symbolic.
Patrick and Goyal \shortcite{patrick2001} applied the AdaBoost
algorithm for boosting the performance of decision graphs.
The latter are an extension of decision trees: they allow tree nodes
to have more than one parent.
The boosting algorithm improves the performance of the decision graphs
by assigning weights to the training data items based on how
accurately they have been classified.
Hammerton \shortcite{hammerton2001b} used a feed-forward neural
network architecture, long short-term memory, for predicting embedded
clause structures.
The network processes sentences word-by-word.
Memory cells in its hidden layer enable it to remember states with
information about the current clause.

D\'ejean \shortcite{dejean2001} predicted clause boundaries with his
symbolic learner ALLiS (Architecture for Learning Linguistic
Structure).
It is based on theory refinement, which means that it adapts grammars.
The learner selects a set of rules based on their prediction accuracy
of classes in a training corpus.
Tjong Kim Sang \shortcite{tks2001b} evaluated a memory-based learner
while using different combinations of features describing items which
needed to be classified.
His learner was well suited for identifying clause starts and clause
ends but less suited for the predicting complete clauses.
Therefore he used heuristic rules for converting the part one and two
results of the shared task to results for the third part.

\footnotetext[2]{Performances on lines with a * suffix are different
from those in the paper version of the CoNLL-2001 proceedings.}
\addtocounter{footnote}{1}

Molina and Pla \shortcite{molina2001} have applied a specialized
Hidden Markov Model (HMM) to the shared task.
They interpreted the three parts of the shared task as tagging problems
and made the HMM find the most probable sequence of tags given an
input sequence.
In the third part of the task they limited the number of possible
output tags and used rules for fixing bracketing problems.
Carreras and M\`arquez \shortcite{carreras2001} converted the clausing
task to a set of binary decisions which they modeled with decision
trees which are combined by AdaBoost.
The system uses features which in some cases contain relevant
information about a complete sentence.
It produces a list of clauses from which the ones with the highest
confidence scores will be presented as output.

We have derived baseline scores for the different parts of the shared
task by evaluating a system that assigns one clause to every sentence.
Each of these clauses completely covers a sentence.
All participating systems perform above the baselines.

In the development data for part 1 of the shared task, at 30 times
all five participating systems (Hammerton's only did part 3 of the
shared task) predicted a clause start at a position where there was
none.
About half of these were in front of the word {\em to}.
The situation in which all five systems missed a clause start
occurred 205 times at positions with different succeeding words.
It seems that many of these errors were caused by a missing
comma immediately before the clause start.

In three cases, the five systems unanimously found an end of a clause 
where there was none in the development data of part 2 of the shared
task.
All these occurred at the end of 'sentences' which consisted of a
single noun phrase or a single adverbial phrase.
In 205 cases all five systems missed a clause end.
These errors often occurred right before punctuation signs.

It is hard to make a similar overview for part 3 of the shared task.
Therefore we have only looked at the accuracies of two clause tags:
(S(S* (starting two clauses) and *S)S) (ending two clauses).
Never did more than three of the six systems correctly predicted
the start of two clauses.
The best performing system for this clause tag was the one of Carreras
and M\`arquez with about 52\% recall.
Three of the systems did not find back any of the double clause starts
and the average recall score of the six was 21\%.
The end of two clauses was correctly predicted by all six systems
about 0.5\% of the times it occurred.
Again, the system of Carreras and M\`arquez was best with 63\%
recall while the average system found back 33\%.

The six result tables show that the system of Carreras and M\`arquez
clearly outperforms the other five systems on all parts of the shared
task.
They were the only one to use input features that contained
information of a complete sentence and it seems that this was a good
choice.

\section{Related Work}

There have been some earlier studies in identifying clauses.
Abney \shortcite{abney90} used a clause filter as a part of his CASS
parser.
It consists of two parts: one for recognizing basic clauses and one 
for repairing difficult cases (clauses without subjects and clauses
with additional VPs).
Ejerhed \shortcite{ejerhed96} showed that a parser can benefit from
automatically identified clause boundaries in discourse.
Papageorgiou \shortcite{papageorgiou97} used a set of hand-crafted
rules for identifying clause boundaries in one text.
Leffa \shortcite{leffa98} wrote a set of clause identification rules
and applied them to a small corpus.
The performance was very good, with recall rates above 90\%.
Or{\u{a}}san \shortcite{oracan2000} used a memory-based learner with
post-processing rules for predicting clause boundaries in Susanne
corpus.
His system obtained F rates of about 85 for this particular task.

\section{Concluding Remarks}

We have presented the CoNLL-2001 shared task: clause identification.
The task was split in three parts: recognizing clause starts, finding
clause ends and identifying complete, possibly embedded, clauses.
Six systems have participated in this shared task.
They used various machine learning techniques,
boosting,
connectionist methods,
decision trees,
memory-based learning,
statistical techniques and
symbolic methods.
On all three parts of the shared task the boosted decision tree system
of Carreras and M\`arquez \shortcite{carreras2001} performed best.
It obtained an F$_{\beta=1}$ rate of 78.63 for the third part of the
shared task.

\section*{Acknowledgements}

We would like to thank SIGNLL for giving us the opportunity to
organize this shared task and our colleagues of
the Seminar f\"ur Sprachwissenschaft in T\"ubingen,
CNTS - Language Technology Group in Antwerp, and 
the ILK group in Tilburg for valuable discussions and comments.
This research has been funded by the European TMR network
Learning Computational Grammars\footnote{http://lcg-www.uia.ac.be/}.

\small

\end{document}